# A New Approach of Improving CFA Image for Digital Camera's

**Manoj Kumar[1], Vikas Kaushik[2], Pradeep Singla[3]**

[1]*Student, Deptt.of Elect.& Comm., Hindu College of Engineering, Sonipat, m.dahiya89@gmail.com*
[2]*Lecturer, Deptt.of Elect.&Comm, SonipatInstt. of Engg.& Mgmt., Sonipat; willingvikas@gmail.com*
[3]*Deptt.of Elect.& Comm., Hindu College of Engineering., Sonipat, pardeep51355@gmail.com*

## ABSTRACT

**This paper work directly towards the improving the quality of the image for the digital cameras and other visual capturing products. In this Paper, the authors clearly defines the problems occurs in the CFA image. A different methodology for removing the noise is discuses in the paper for color correction and color balancing of the image. At the same time, the authors also proposed a new methodology of providing denoisiing process before the demosaickingfor the improving the image quality of CFA which is much efficient then the other previous defined. The demosaicking process for producing the colors in the image in a best way is also discuss.**

## I. INTRODUCTION

In the time of embedded system, the digital camera's are one of the popular consumer electronic product. The personal digital assistance (PDA's), mobile cell phones, iPods are embedded with the expansive digital camera's instead of film camera's for capturing or recording of the activities of everyday life. The removing of noise or providing a correction of non-linearity's of sensor of camera'snon uniformities, adjusting the white balance and many more needs a significant processing for users viewable image [3]. These camera's are generally uses a sensor with CFA (color Filter Array) which is a very important part of processing chain. Any color image is consisting of three basic primary color R, G, B. the only one third or single color is to be measured at each pixel by CFA and the remaining missing true color image is estimated by camera and this estimated process by the camera is known as demosaking [1-5].

### I.1 Motivation behind work

A digital image is an array of real or complex numbers represented by a finite numbers of bits. Any imageconsists of picture-element called pixels or pel [14]. An image acquired by optical, electro-optical means is likely to be degraded by the sensing environment. The degradation may be in form of sensor noise, blur due to camera miss focus, relative object camera motion, random atmospheric turbulence and so on. So, now a days denoising of image is main aspect to improve image quality by modifying or changing image parameter such as light luminance, brightness and contrast etc[16]. These digital image denoising has a broad spectrum of applications such as remote sensing via satellite & other spacecraft's, image transmission & storage for business applications, medical processing, radar and acoustic image processing, robotics and automated inspection of industrial parts.

## II. BACKGROUND OF CFA

The digital cameras uses a very precious part i.e., single sensor with a colour filter array (CFA) for capturing the visual scene in color form as shown in fig.1.

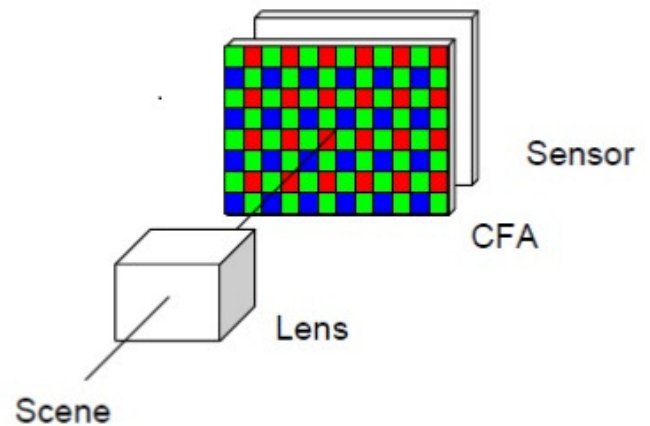

**Fig.1- Demonstration of optical path digital camera**

As we have discussed in the last section the sensor cell can record only one colour value.The other two missingcolour components at each position need to be interpolated from the available CFA sensor readings to reconstruct the full colour image. The colour interpolation process is usually called colourdemosaicing (CDM).

There are many patterns out of which a CFA can have any pattern. The most commonly used CFA pattern is Bayer pattern shown in fig. 2. A Bayer filter mosaic is a color filter array (CFA) for arranging RGB color filters on a square grid of photo sensors. Its particular arrangement of color filters is used in most single-chip digital image sensors used in digital cameras, camcorders, and scanners to create a color image. The filter pattern is 50% green, 25% red and 25% blue, hence is also called RGBG, GRGB, or RGGB. The Bayer array measures the G image on a quincunx grid an the R & B images on rectangular grids. The G image is measured at higher sampling rate because sensitivity of human eyelie in medium wavelengths, corresponding to the G portion of the spectrum.

| G | B | G | B | G | B | G | B |
|---|---|---|---|---|---|---|---|
| R | G | R | G | R | G | R | G |
| G | B | G | B | G | B | G | B |
| R | G | R | G | R | G | R | G |
| G | B | G | B | G | B | G | B |
| R | G | R | G | R | G | R | G |
| G | B | G | B | G | B | G | B |
| R | G | R | G | R | G | R | G |

**FIG.2. Bayer patterned CFA.**

There are number of applications where noise is present in the CFA. The presence of noise in CFA data not only deteriorates the visual quality of captured images but also often cause serious demos icing artifacts which can be extremely difficult to remove using a subsequent denoising process. This problem can be illustrated by taking an example shown in fig.3. where we use the noise free image as an example.

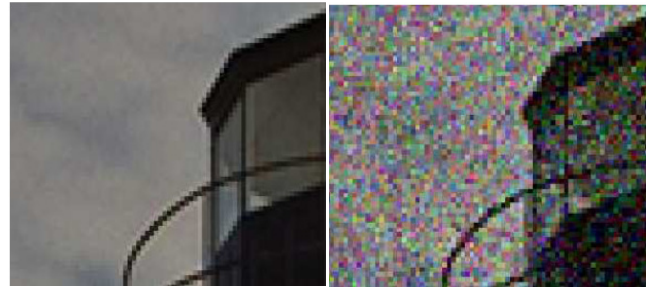

**Fig. 3(a) noise free image Fig. 3(b) Noisy image**

In these graphics we can see that in fig.3(b) there are some noise added in the graphics due to surounding effects or due to some device capabilities after capturing. So, for the user's interest, there is a need of removing such kind of noises. There are different methods proposed in the literature used by the different scientist/researchers discussed in the next section.

### III. METHODOLOGY FOR DENOISING

Many CDM algorithm [1]-[8] proposed in the past are based on unrealistic assumptions of noise free CFA data.To suppress the effect ofnoise on the demosaicked image, three strategies are possible.

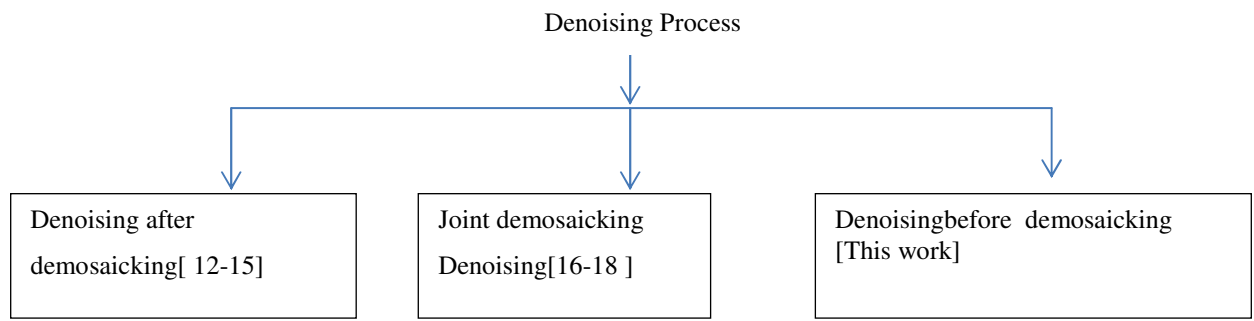

**Fig.4 Demonstration of different denoising process**

### III.1 Denoising after demosaicking

A convenient strategy to remove noise is to denoise the demosaicked images. Algorithms developed for gray-scale imaging, for example [12]–[15], can be applied to each channel of the demosaicked color image separately whereas some color image filtering techniques [11] process color pixels as vectors. The problem of this strategy is that noisy sensor readings are roots of many color artifacts in demosaicked images and those artifacts are difficult to remove by denoising the demosaicked full-color data. In general the CFA readings corresponding to different color components have different noise statistics. The CDM process blends the noise contributions across channels, thus producing compoundnoise that is difficult to characterize. This makes the design of denoising algorithms for single-sensor color imaging very difficult as demonstrated below.

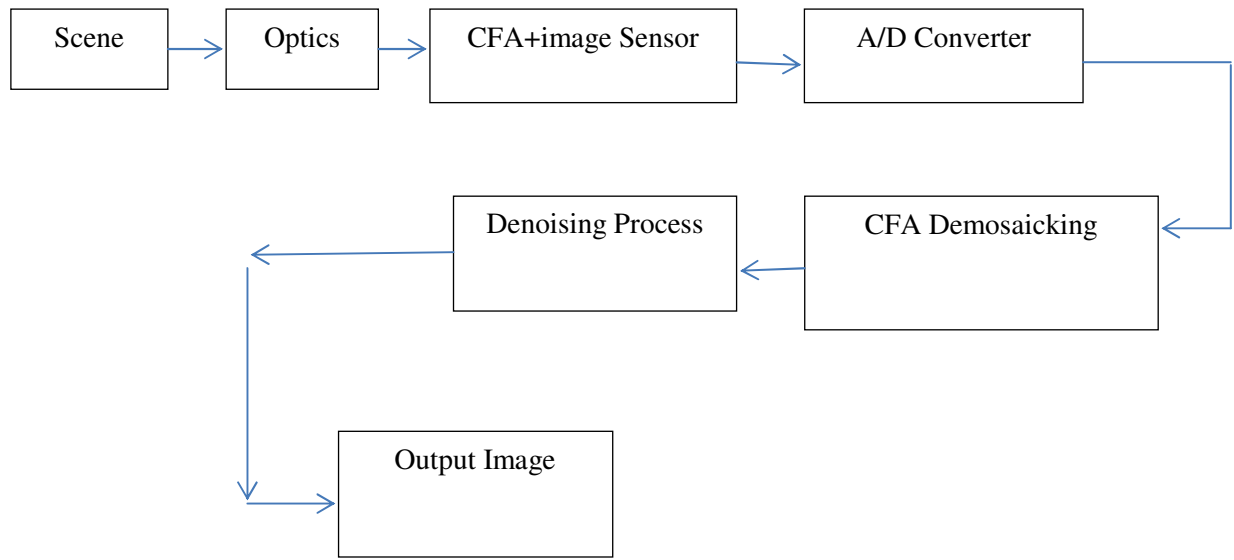

**Fig.5: Demonstration of Denoising after demosaicking**

### III.2 Joint demosaicking-denoising

Recently, some schemes that perform demosaicking and denoising jointly have been proposed [16]–[18]. In [17], Trussell and Hartig presented a mathematical model for color demosaicking using minimum mean square error (MMSE) estimator. The additive white noise is considered in the modeling. Ramanath and Snyder [20] proposed a bilateral filter based demosaicking method. Since bilateral filtering exploits the similarity in both spatial and intensity spaces, this scheme can handle light noise corrupted in the CFA image. Hirakawa and Parks [4] developed a joint demosaicking-denoising algorithm by using the total least square (TLS) technique where both demosaicking and denoising are treated as an estimation problem with the estimates being produced from the available neighboring pixels.The filter used for joint demosaicking-denoising is determinedadaptively using the TLS technique under some constraints of the CFA pattern. The joint demosaicking-denoising scheme developed by Zhang *et al.* [13] first performs demosaicking-denoising on the green channel. The restored green channel is then used to estimate the noise statistics in order to restore the red and blue channels. In implementing the algorithm, Zhang *etal.*estimated the red-green and blue-green color difference images rather than directly recovering the missing color samples by using a linear model of the color difference signals. Inspired by the directional linear minimum mean square-error estimation (DLMMSE) based CDM scheme in proposed an effective nonlinear and spatially adaptive filter by using local polynomial approximation to remove the demosaicking noise generated in the CDM process and then adapted this scheme to noisy CFA inputs for joint demosaicking-denoising.

### III.3 Denoising before demosaicking (Proposed Method)

The third way to remove noise from CFA data is to implement denoising before demosaicking which is our proposed method. However, due to the underlying mosaic structure of CFAs, many existing effective monochromatic image denoising methods cannot be applied to the CFA data directly. To overcome the problem, the CFA image can be divided into several sub-images using the approach known from the CFA image compression literature, e.g.[19].

Since each of the sub-images constitutes a gray-scale image, it can be enhanced using denoising algorithms from gray-scale imaging. The desired CFA image is obtained by restoring it from the enhanced sub-images. Nonetheless, such a scheme does not exploit theinterchannel correlation which is essential to reduce various color shifts and artifacts in the final image [11]. Since the volume of CFA images is three times less than that of the demosaicked images, there is a demand to develop new denoising algorithms which can fully exploit the interchannel correlations and operate directly on CFA images, thus achieving higher processing rates.

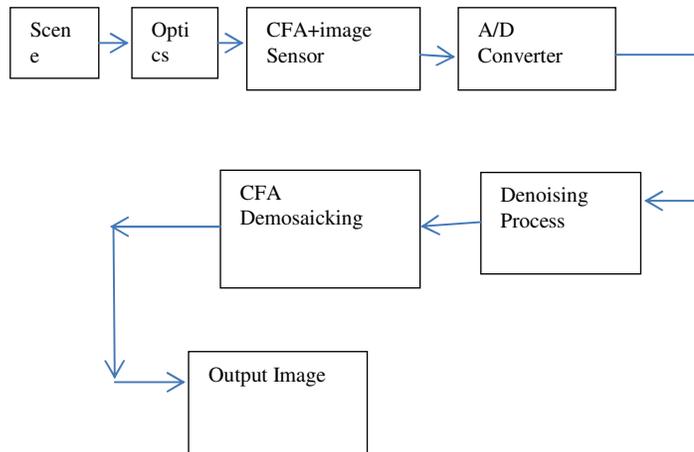

### III.4 Conclusion and Discussion

In this paper, the authors providean efficient new method of removing the noises in the CFA data by providing the denoising operation before the demosaicking. The different methods for improving the color balancing and color correction of CFA data has been also discussed by the authors. The complete information of noise produced in the Bayer filter has been explained. The new proposed method of denoisingis also a better option and efficient option for the color improvement which is proposed in this paper. This proposed method is also a better option in the field of image processing.

## REFERENCES

[1] P. Longère, X. Zhang, P. B. Delahunt, and D. H. Brainard, "Perceptual assessment of demosaicing algorithm performance," *Proc. IEEE*, vol. 90, no. 1, pp. 123–132, Jan. 2002.

[2] B. K. Gunturk, Y. Altunbasak, and R. M. Mersereau, "Color plane interpolation using alternating projections," *IEEE Trans. Image Process.*, vol. 11, no. 9, pp. 997–1013, Sep. 2002.[

[3] B. K. Gunturk, J. Glotzbach, Y. Altunbasak, R. W. Schafer, and R. M. Mersereau, "Demosaicking: Color filter array interpolation in singlechip digital cameras," *IEEE Signal Process. Mag.*, vol. 22, no. 1, pp. 44–54, Jan. 2005.

[4] K. Hirakawa and T. W. Parks, "Adaptive homogeneity-directed demosaicing algorithm," *IEEE Trans. Image Process.*, vol. 14, no. 3, pp. 360–369, Mar. 2005.

[5] L. Zhang and X. Wu, "Color demosaicking via directional linear minimum mean square-error estimation," *IEEE Trans. Image Process.*, vol.14, no. 12, pp. 2167–2178, Dec. 2005.

[6] R. Lukac, K. Martin, and K. N. Plataniotis, "Demosaicked image postprocessing using local color ratios," *IEEE Trans. Circuits Syst. VideoTechnol.*, vol. 14, no. 6, pp. 914–920, Jun. 2004.

[7] X. Li, "Demosaicing by successive approximation," *IEEE Trans. Image Process.*, vol. 14, no. 3, pp. 370–379, Mar. 2005.

[8] D. D. Muresan and T. W. Parks, "Demosaicing using optimal recovery," *IEEE Trans. Image Process.*, vol. 14, no. 2, pp. 267–278, Feb. 2005.

[9] R. Lukac and K. N. Plataniotis, "Color filter arrays: design and performance analysis," *IEEE Trans. Consum. Electron.*, vol. 51, no. 11, pp. 1260–1267, Nov. 2005.

[10] D. Alleysson, S. Susstrunk, and J. Herault, "Linear demosaicing inspired by the human visual system," *IEEE Trans. Image Process.*, vol. 14, no. 4, pp. 439–449, Apr. 2005.

[11] R. Lukac, B. Smolka, K. Martin, K. N. Plataniotis, and A. N. Venetsanopoulos, "Vector filtering for color imaging," *IEEE Signal Process. Mag.*, vol. 22, no. 1, pp. 74–86, Jan. 2005.

[12] S. G. Chang, B. Yu, and M. Vetterli, "Spatially adaptive wavelet thresholding with context modeling for image denoising," *IEEE Trans. Image Process.*, vol. 9, no. 9, pp. 1522–1531, Sep. 2000

[13] L. Zhang, P. Bao, and X. Wu, "Multiscale LMMSE-based image denoising with optimal wavelet selection," *IEEE Trans. Circuits Syst.Video Technol.*, vol. 15, no. 4, pp. 469–481, Apr. 2005.

[14] J. Portilla, V. Strela, M. J. Wainwright, and E. P. Simoncelli, "Image denoising using scale mixtures of gaussians in the wavelet domain," *IEEE Trans. Image Process.*, vol. 12, no. 11, pp. 1338–1351, Nov. 2003.

[15] A. Pizurica and W. Philips, "Estimating the probability of the presence of a signal of interest in multiresolution single- and multiband image denoising," *IEEE Trans. Image Process.*, vol. 15, no. 3, pp. 654–665, Mar. 2006

[16] L. Zhang, X. Wu, and D. Zhang, "Color reproduction from noisy CFA data of single sensor digital cameras," *IEEE Trans. Image Process.*, vol. 16, no. 9, pp. 2184–2197, Sep. 2007.

[17] H. J. Trussell and R. E. Hartwig, "Mathematics for demosaicking," *IEEE Trans. Image Process.*, vol. 11, no. 4, pp. 485–492, Apr. 2002. 18. H. J. Trussell and R. E. Hartwig, "Mathematics for demosaicking," *IEEE Trans. Image Process.*, vol. 11, no. 4, pp. 485–492, Apr. 2002.

[18] C.C. Koh, J. Mukherjee, and S. K. Mitra, "New efficient methods of image compression in digital cameras with color filter array," *IEEE Trans. Consum. Electron.*, vol. 49, no. 11, pp. 1448–1456, Nov. 2003

[19] Rajeev Ramanath and Snyder "Adaptivedemosaking" Journel of electronic imaging" 633-642, Oct 2003.